\documentclass[conference]{IEEEtran}
\IEEEoverridecommandlockouts

\usepackage{cite}
\usepackage{amsmath,amssymb,amsfonts}
\usepackage{algorithmic}
\usepackage{graphicx}
\usepackage{textcomp}
\usepackage{xcolor}
\usepackage{amsmath}
\usepackage[autostyle=true]{csquotes}
\usepackage[nolist]{acronym}
\def\IEEEkeywordsname{Keywords}

\begin{document}

\begin{acronym}[ECU]
\acro{cnn}[CNN]{Convolutional Neural Network}
\acroplural{cnn}[CNNs]{Convolutional Neural Networks}
\acro{fpga}[FPGA]{Field Programmable Gate Array}
\acroplural{fpga}[FPGAs]{Field Programmable Gate Arrays}
\acro{rcnn}[RCNN]{Region-based Convolutional Neural Network}
\acroplural{rcnn}[RCNNs]{Region-based Convolutional Neural Networks}
\acro{ssd}[SSD]{Single Shot Detector}
\acro{yolo}[YOLO]{You Only Look Once}
\acro{rms}[RMSprop]{Root Mean Square Propagation}
\acro{adam}[ADAM]{Adaptive Moment Estimation}
\acro{flop}[FLOP]{Floating Point Operation}
\acroplural{flop}[FLOPS]{Floating Point Operations}
\acro{map}[MAP]{Mean Average Precision}
\acro{relu}[ReLU]{Rectified Linear Unit}
\acro{lrelu}[ReLU$_\alpha$]{Leaky Rectified Linear Unit}
\acro{mse}[MSE]{Mean Squared Error}
\acroplural{mse}[MSEs]{Mean Squared Errors}
\acro{vgg}[VGG]{Visual Geometry Group}
\acro{bbox}[bbox]{Bounding Box}
\end{acronym}

\title{An Overview of Arithmetic Adaptations for Inference of Convolutional Neural Networks on Re-configurable Hardware \\
}

\author{\IEEEauthorblockN{Ilkay Wunderlich}
\IEEEauthorblockA{\textit{Institute of Computer Engineering} \\
\textit{Technische Universit\"at Dresden}\\
Dresden, Germany \\
email: ilkay.wunderlich@tu-dresden.de}
\and
\IEEEauthorblockN{Benjamin Koch}
\IEEEauthorblockA{\textit{AVI Systems} \\
Freital, Germany \\
email: benjamin.koch@avi-systems.eu}
\and
\IEEEauthorblockN{Sven Sch\"onfeld}
\IEEEauthorblockA{\textit{AVI Systems} \\
Freital, Germany \\
email: sven.schoenfeld@avi-systems.eu}
}

\maketitle

\begin{abstract}
\acp{cnn} have gained high popularity as a tool for computer vision tasks and for that reason are used in various applications.
There are many different concepts, like single shot detectors, that have been published for detecting objects in images or video streams. 
However, \acp{cnn} suffer from disadvantages regarding the deployment on embedded platforms such as re-configurable hardware like \acp{fpga}. Due to the high computational intensity, memory requirements and arithmetic conditions, a variety of strategies for running \acp{cnn} on \acp{fpga} have been developed.
The following methods showcase our best practice approaches for a TinyYOLOv3 detector network on a XILINX Artix-7 FPGA using techniques like fusion of batch normalization, filter pruning and post training network quantization.\\
\end{abstract}


\section{Introduction}
This section introduces the historic background of \acp{ssd} and the challenges of implementing \acp{cnn} on reconfigurable hardware. Afterwards, the general \enquote{life} of neural networks is explained and expanded with the adaptation stage.
\subsection{Single shot detector network}
After the success of \acfp{cnn} for image classification, object detection stepped into the focus of research. A first brute force approach used sliding windows throughout the image with a classification network. This strategy limits itself to the granularity of the window size and window strides.\\ \indent
With \acp{rcnn} a more sophisticated tool for object detection was presented. The \ac{rcnn} itself contains two \acp{cnn}, which are solving the tasks of detecting objects of interest and classifying the found objects \cite{i01}.\\ \indent
The first \ac{ssd} was published by Joseph Redmon et al. with the iconic name \ac{yolo} \cite{i02}\cite{i03}. The first version was supplemented by two updates: YOLO9000, which is also known as YOLOv2 \cite{i04} and the most recently YOLOv3 \cite{i05}. Additionally to the \ac{yolo} versions smaller versions, named TinyYOLOvX, are provided. The main focus for the following elaborations is put on the TinyYOLOv3 architecture, which optimises the trade off between detection performance and computational effort.

\subsection{Challenges on re-configurable hardware}
Implementing \ac{cnn} on re-configurable hardware introduces several constraints, that affect the architecture of the used networks as well as the underlying arithmetic operations, memory access and scheduling of operations.\\ \indent
Using state of the art \acfp{fpga}, e.g., XILINX Artix-7, only fixed-point arithmetic can be implemented in an efficient way, so quantization of the network model will be necessary. Because of that, choosing quantization factors and a proper format for intermediate results to keep the deviation regarding the fixed-point model as low as possible is a key consideration to adapt a model for hardware inference.\\ \indent
The most challenging constraint is the limited number of logic elements to implement the building blocks of the \ac{cnn}. Dissecting the parts of the neural network in candidates for hardware and software implementation is a key consideration to be made in the system architecture. Since \acp{fpga} typically have little on-chip memory an efficient way to access memory has to be part of the architecture design as well. 
Also, the typical lower clock frequency in re-configurable hardware adds another trade-off.\\ \indent
This leads to the conclusion that a general purpose accelerator is hard to design. Every use case and neural network should be analysed according to needed performance and target platform.

\subsection{Adaptation Stage}
The \enquote{life} of any neural network can be categorized into two stages: training and inference stage. At the training stage the neural network gets trained for its later task using labelled data, which gets divided into training and test sets. A variety of optimization techniques exists for the training stage. For example, one common target of optimizing the training stage aims on reducing the amount of needed iterations to reach the global minimum of the loss function. Several machine learning libraries implement optimizers for speeding up training, like \ac{rms} \cite{i1} or \ac{adam} \cite{i2}.\\ \indent
The inference stage is the application of the neural network, after it is properly trained. Adaptations to the network might be needed, depending on the hardware platform on which the inference stage takes place. The entirety of the adaptation work flow is summarized as the adaptation stage.\\ \indent
In section \ref{sec_bn} and \ref{sec_fp}, two optional but very useful adaptation steps are introduced. In Section \ref{sec_quant}, the mandatory adaptation of switching the arithmetic backbone of the network from floating point to fix point operations is presented. The benefits of these techniques are summarized in section \ref{sec_res}.

\section{Batchnorm Fusing}
In this section, the concept of batch normalization and the general layout of a convolution layer are briefly explained. Thereafter, formulae and results of eliminating the batch normalization layer are given.
\label{sec_bn}
\subsection{Background of Batch Normalization}
\begin{figure}[htbp]
\centerline{\includegraphics[width=0.5\textwidth]{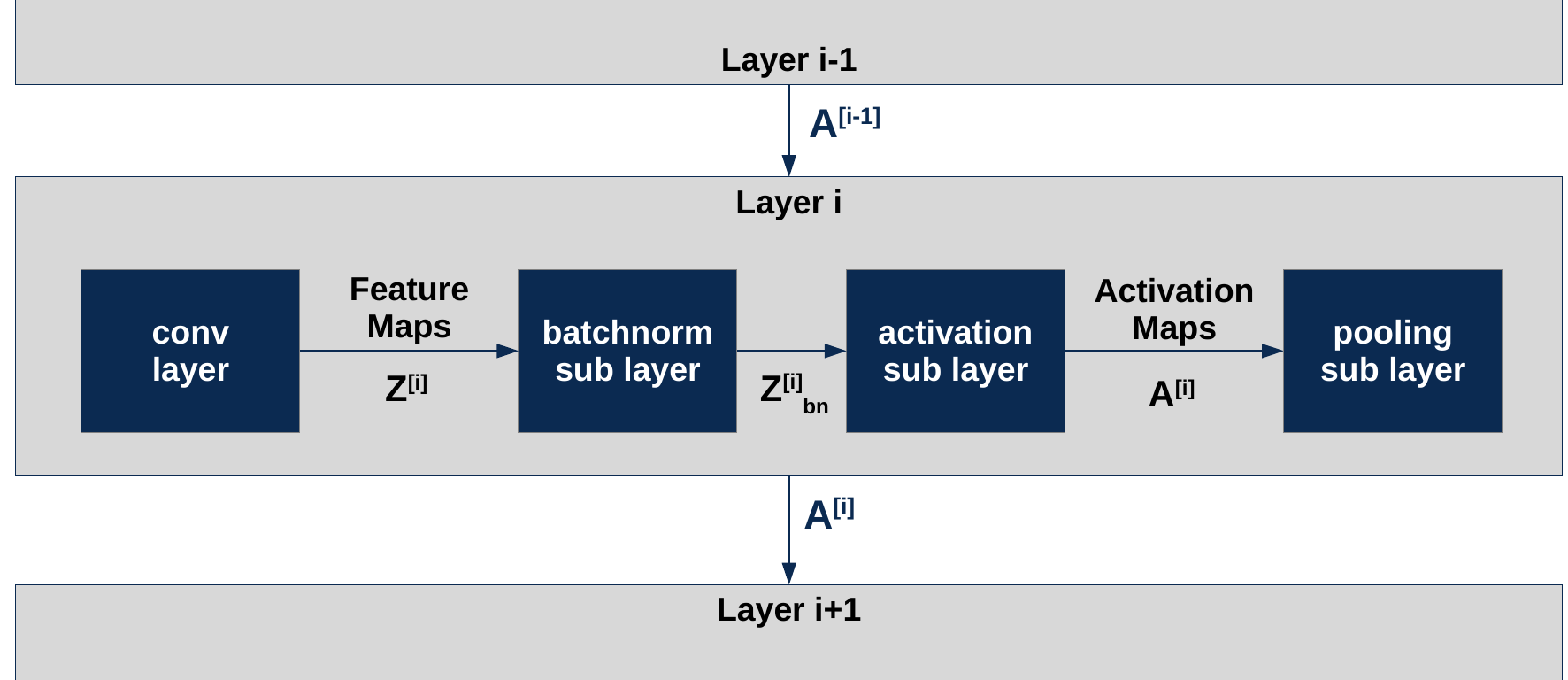}}
\caption{Example of a general convolutional layer (abbreviated with conv) with its subsequent batchnorm, activation and pooling sub layer and their output descriptions.}
\label{fig_bn1}
\end{figure}

Batch normalization \cite{b1}, which is  often times abbreviated with batchnorm, is a sub layer used to reduce internal covariate shifts. These shifts are defined as changes in the distribution of the network's activation, which are caused by changes in the parameters of the network during training stage. Diminishing the covariate shift enables higher learning rates, reduces the risk of getting stuck in poor local minima and prevents vanishing or exploding gradients. Another advantageous side effect of batch normalization is the increased generalization ability of the network \cite{b1}. Many modern networks are using batchnorm sub layers, e.g., YOLOv3 \cite{i05}, MobileNet \cite{b3} and ResNet \cite{b4}.\\ \indent
The batch normalization sub layer is located between the convolutional layer and the activation sub layer which is illustrated for layer $i$ in Figure \ref{fig_bn1}. 
\\ \indent
This sub layer contains a set of up to four parameters:
\begin{enumerate}
\item \textbf{Mini-Batch Mean}:\\
The mini-batch mean $\mu^{[i]}$ is measured regarding the mean of the feature maps of the current mini-batch $\mathcal{B}^{[i]}=\left\lbrace z^{[i]}_1,...,z^{[i]}_m\right\rbrace$ at each batchnorm sub layer $i$ in the network. This metric gets updated for each mini-batch and epoch during the training stage:
\begin{equation}
\mu^{[i]} \leftarrow \frac{1}{m} \cdot \sum_{j=0}^{m} \mathcal{B}^{[i]}
\end{equation}
\item \textbf{Mini-Batch Variance}:\\
Similar to mini-batch mean with respect to the variance of the current mini-batch:
\begin{equation}
{\sigma^2}^{[i]} \leftarrow \frac{1}{m} \cdot \sum_{j=0}^{m} (\mathcal{B}^{[i]}-\mu^{[i]})^2
\end{equation}
\item \textbf{Scale}:\\
Trained scaling term $\gamma^{[i]}$, which is an optional and trainable component of the batchnorm sub layer. 
\item \textbf{Shift}:\\
Trained shifting term $\beta^{[i]}$, which is an optional and trainable component of the batchnorm sub layer as well. Commonly used instead of bias parameters $b$ in the convolutional layer.
\end{enumerate}
The batchnorm parameters are one dimensional vectors, which are applied to each feature map from the previous convolutional layer separately. The amount of elements is determined by the amount of filters used in the conv layer. The forward propagation step for the batchnorm sub layer is given in \eqref{eq_bn1}:
\begin{equation}
Z_{bn}^{[i]}= BN(Z^{[i]})= \gamma^{[i]} \cdot \underbrace{\frac{Z^{[i]}-\mu^{[i]}}{\sqrt{{\sigma^2}^{[i]}+\epsilon}}}_{normalization} + \beta^{[i]}
\label{eq_bn1}
\end{equation}
where $\epsilon$ is a small scalar value added to the variance to provide numerical stability (e.g., keras default: $\epsilon=0.001$ \cite{b5}). 

\subsection{Fusion of Batchnorm Parameters into Convolutional Parameters}
In order to get rid of the computational effort of the batch normalization sub layer, it is recommended to fuse the batchnorm parameters into the convolutional parameters before entering the inference stage. To derive the formulae for batchnorm fusing, the kernel convolution, which is performed filter-wise in the convolutional layer is introduced for layer $i$:
\begin{equation}
Z^{[i]} = Conv\left(A^{[i-1]};W^{[i]},b^{[i]}\right) = A^{[i-1]} \ast W^{[i]} + b^{[i]}
\label{eq_bn2}
\end{equation} 
where $A^{[i-1]}$ denotes the three dimensional activation map from the previous layer of matrix shape $(w,h,c)$ - (width,height,color channels). 
$W^{[i]}$ is the filter matrix of shape $(f_w,f_h,c)$ and $b^{[i]}$ represents the optional bias vector is shape $(c,1)$. 
For easier reading, the layer indices $()^{[i]}$ are skipped in the following arguments with hinting $A^{[i-1]}$ as $A_{prev}$.
It can be shown, that the convolution operation $Conv(A;W,b)$ holds the following property:
\begin{align}
k \cdot Conv\left(A_{prev};W,b\right)+h = Conv\left(A_{prev};k \cdot W,k \cdot b+h\right)
\label{eq_bn3}
\end{align}
for $k,h=const.$ and $k,h \in \mathbb{R}^c$. With \eqref{eq_bn1},\eqref{eq_bn2} and \eqref{eq_bn3} the formulae for batchnorm fusing are derived as follows:\\[0.5em]
Replacing $Z^{[i]}$ from \eqref{eq_bn1} with \eqref{eq_bn2}:
\begin{equation}
Z_{bn} = \gamma \cdot \frac{Conv\left(A_{prev};W,b\right)-\mu}{\sqrt{{\sigma^2}+\epsilon}}+\beta
\end{equation}
Reordering to get the equation above to a similar form as shown in \eqref{eq_bn3}:
\begin{equation}
Z_{bn} = \underbrace{\frac{\gamma}{\sqrt{{\sigma^2}+\epsilon}}}_{k} \cdot Conv\left(A_{prev};W,b\right) + \underbrace{\beta - \frac{\gamma \cdot \mu}{\sqrt{{\sigma^2}+\epsilon}}}_{h}
\end{equation}
which can be written as:
\begin{equation}
{\textstyle 
Z_{bn} = Conv\left(A_{prev};\frac{\gamma}{\sqrt{{\sigma^2}+\epsilon}} \cdot W, \frac{\gamma}{\sqrt{{\sigma^2}+\epsilon}} \cdot b + \beta - \frac{\gamma \cdot \mu}{\sqrt{{\sigma^2}+\epsilon}}\right)=}\label{eq_bn4}
\end{equation}
\begin{equation*}
Conv\left(A_{prev};\frac{\gamma}{\sqrt{{\sigma^2}+\epsilon}} \cdot W, \frac{\gamma}{\sqrt{{\sigma^2}+\epsilon}} \cdot (b-\mu) + \beta\right)
\end{equation*}
With \eqref{eq_bn4} the formulae for the fused parameters $W_{bn}$, $b_{bn}$ are derived as:
\begin{equation}
W_{bn}=\frac{\gamma}{\sqrt{{\sigma^2}+\epsilon}} \cdot W
\end{equation}
\begin{equation}
b_{bn}=\frac{\gamma}{\sqrt{{\sigma^2}+\epsilon}} \cdot (b-\mu) + \beta 
\label{eq_bn5}
\end{equation}
\begin{equation}
Z_{bn}=Conv\left(A_{prev};W_{bn},b_{bn}\right)
\end{equation}
For convolution layers trained without biases \eqref{eq_bn5} is reduced to:
\begin{equation}
b_{bn}= \beta  - \frac{\gamma \cdot \mu}{\sqrt{{\sigma^2}+\epsilon}} 
\end{equation}

\subsection{Benefits of Batchnorm fusing}
\label{sec_bn_benefit}
\begin{figure}[htbp]
\centerline{\includegraphics[width=0.5\textwidth, trim = 5 0 5 0, clip]{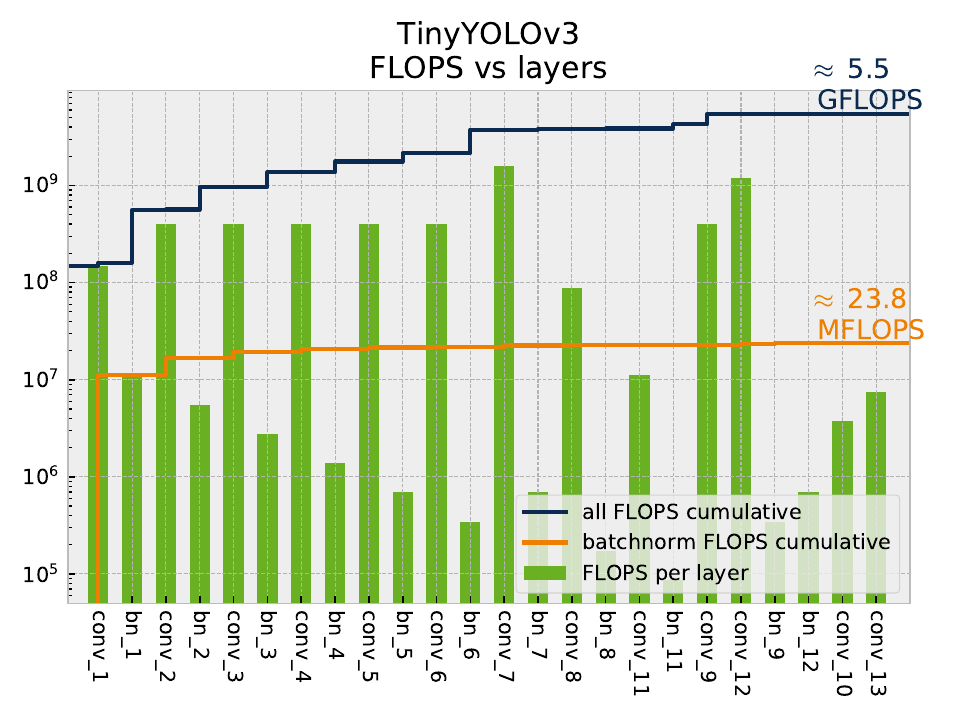}}
\caption{Required \acp{flop} (y-axis, logarithmic) for all convolutional layer with their batchnorm sub layer (x-axis) for the TinyYOLOv3 architecture.}
\label{fig_bn2}
\end{figure}

The emphasis for estimating the benefits of batchnorm fusing is shown with respect to the reduction of \acp{flop}. In Figure \ref{fig_bn2}, the amount of needed \acp{flop} for the first layers of a TinyYOLOv3 network without batchnorm fusing for an input image size of $416\times416\times3$ (width, height, color channels) is shown. This network requires approximately $5.5$ G\acp{flop} for processing one image from the input layer to the output layers, excluding the processing steps in the \ac{yolo} back-end. After fusing the batchnorm parameters into the convolutional parameters, the FLOPS count is reduced by $23.8$ M\acp{flop}.
\\ \indent
This decrease sounds low with a reduction factor of only $0.4\%$. 
But more important is the avoidance of batchnorm sub layers as a whole, because each sub layer requires additional logic elements, complexity in the control loop and power.
Another benefit is the preparation for pruning, which is performed on the fused weight matrices.

\section{Filter Pruning}
The general concept of pruning and the proposed routine are presented. The results of pruning are displayed using parameter count and \acp{flop} as optimization target.
\label{sec_fp}
\subsection{Pruning Background}
In contrast to batch normalization being introduced in 2015, research on pruning of neural networks already started at the end of 1980s \cite{p1} and the beginning of 1990s \cite{p2}. The goal of pruning is reducing the network size by removing redundant connections while maintaining the performance of the network. M.C. Mozer and P. Smolensky pictorially call this technique as \enquote{trimming the fat from a network} \cite{p2}. In order to determine redundant connections several metrics are proposed in literature \cite{p3}. In the following, the emphasis is put on \acp{cnn} and particularly on pruning whole filters after batchnorm fusing. 

\subsection{Pruning Metrics}
In this section, two metrics for determining filter candidates suitable for pruning are introduced. These metrics help finding filters $f_n$ of the weight matrix $W$, which have no or low impact for the forward propagation. The total amount of filters stored in the weight matrix $W$ is denoted with $nf$.
A trivial example for a prunable filter is one which coefficients are all zero ($f_n=\textbf{0}$).
\begin{enumerate}
\item The \textbf{\textsc{Frobenius} Norm} for filter $f_n$ is defined as the following scalar value:
\begin{equation}
\left|\left| f_n \right|\right|_F = \sqrt{\sum_{w,h,c} \left(f_n[w,h,c]\right)^2} 
\label{Frob_1}
\end{equation}
This definition gets expanded for the whole weight matrix by stacking the norms of the all filters $f_n$ to one vector:
\begin{equation}
\left|\left| W \right|\right|_F = \left[ \left|\left|f_0 \right|\right|_F, ..., \left|\left|f_n \right|\right|_F, ... , \left|\left|f_{nf}\right|\right|_F \right]^T \label{Frob_2}
\end{equation}
\item The \textbf{filter sparsity} is a metric for determining the sparsity of a filter. It is defined as the percentage of values close to zero of a filter $f_n$:
\begin{equation}
Sp_\epsilon(f_n) = 1 - \frac{\mathcal{C}(|f_n|<\epsilon)}{\mathcal{C}(f_n)}
\label{Sp_1}
\end{equation}
where $\mathcal{C}(f_n)$ denotes the cardinality of the filter $f_n$ \eqref{Sp_1_help1} and $\mathcal{C}(|f_n|<\epsilon)$ the conditional cardinality of $f_n$ \eqref{Sp_1_help2}.
\begin{equation}
\mathcal{C}(f_n) = \sum_{w,h,c} 1
\label{Sp_1_help1}
\end{equation}
\begin{equation}
\mathcal{C}(|f_n|<\epsilon) = \sum_{w,h,c}  \left\{\begin{array}{ll} 1, & |f[w,h,c]|<\epsilon \\ 0 , & |f[w,h,c]| \geq \epsilon  \end{array}\right. 
\label{Sp_1_help2}
\end{equation}
Similarly to \eqref{Frob_2} this equation is expanded to:
\begin{equation}
Sp_\epsilon(W) = \left[ Sp_\epsilon(f_0) , ...,Sp_\epsilon(f_n), ... , Sp_\epsilon(f_{nf})\right]^T  
\label{Sp_2}
\end{equation}
\end{enumerate}

\subsection{Pruning Routine}
The previously mentioned metrics are used in the proposed pruning routine, which is based on the following inputs:
\begin{itemize}
\item The maximum deviation of \ac{map}: $\Delta_{MAP}$.
\item A representative pruning data set to continuously calculate the mean average precision.
\item An optional Starting threshold $T_{start}$, which is $0$ by default.
\item A value by which the threshold gets incremented $\delta_T$: \\E.g., $\delta_T=0.01$.
\end{itemize}
The pruning routine determines the chosen metric for every filter in the \ac{cnn}, as well as the initial \ac{map} of the network on the pruning data set beforehand. Afterwards every filter, which is below the threshold $T$, is removed and the \ac{map} is calculated again. As long as deviation is lower than the maximum deviation $\Delta_{MAP}$, the threshold is incremented by $\delta_T$. This procedure is repeated until $\Delta_{MAP}$ is reached. \\ \indent
After the pruning routine is finished, the possibility of additional training in order to slightly fine tune the remaining filters to reach the initial \ac{map} is possible. It is advisable to perform the fine tuning with a low learning rate and early stopping to avoid over fitting of the network to the pruning data set.

\subsection{Pruning Result} \label{sec_pruning_results}
\begin{figure}[htbp]
\centerline{\includegraphics[width=0.5\textwidth , trim = 0 10 0 10, clip]{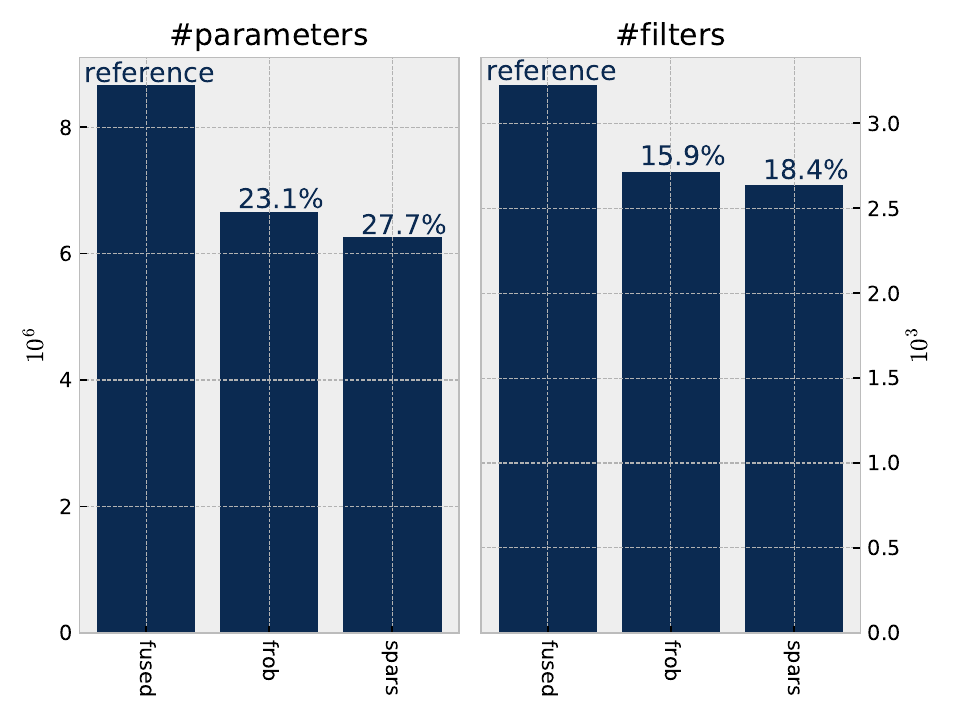}}
\caption{Pruning result showing for the TinyYOLOv3 network example comparing fused model (fused) and pruned model with \textsc{Frobenius} Norm (fro) and Filter Sparsity (spars) as metric. Top of bar: reduction percentage or reference to it. Left: total parameter count. Right: filter count of the network.}
\label{fig_pru1}
\end{figure}

\begin{figure}[htbp]
\centerline{\includegraphics[width=0.5\textwidth , trim = 5 150 5 0, clip]{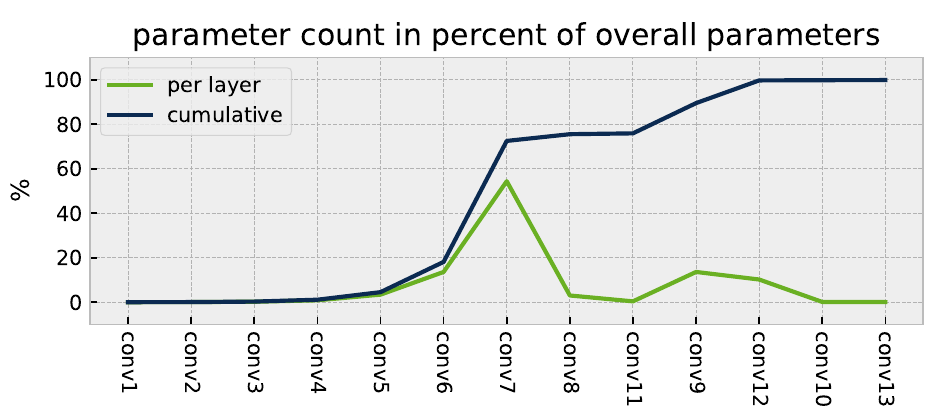}}
\caption{Percentage of parameters for each layer. Per layer and cumulative.}
\label{fig_pru2}
\end{figure}

\begin{figure}[htbp]
\centerline{\includegraphics[width=0.45\textwidth, trim = 0 10 0 10, clip]{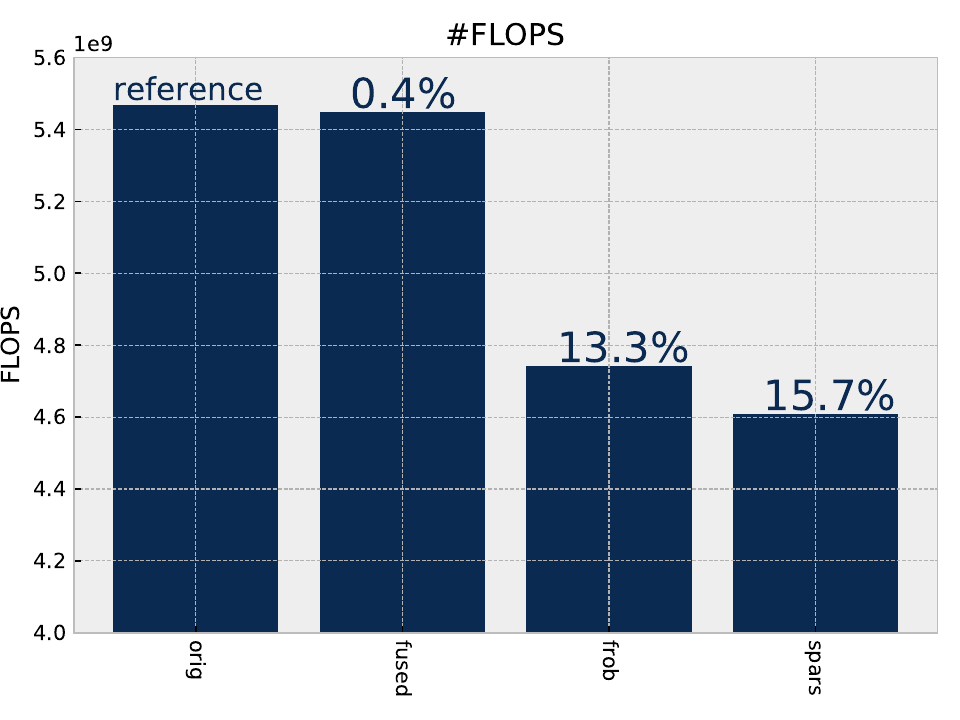}}
\caption{\acp{flop} comparison (y-axis) for the TinyYOLOv3 network example. Reduction percentages (top of bar) using the \enquote{unfused} original model (orig) as reference.}
\label{fig_pru3}
\end{figure}

The following example is based on a TinyYOLOv3 network, which is trained on the TU Darmstadt Pedestrian Dataset \cite{p4}. 
The original model, fused model, and the pruned example models as well as additional information are provided in a separate GitHub repository \cite{p5}.\\ \indent
The maximum deviation of \ac{map} is set to $\Delta_{mAP}=1\%$ and the threshold increment $\delta_T=0.02$. 
In Figure \ref{fig_pru1}, the results of the pruning routine are shown using \textsc{Frobenius} Norm $\left|\left| W \right|\right|_F$ and filter sparsity $Sp_{\epsilon=0.003}(f_n)$ as the pruning metric. 
The results cover the total parameter and filter count of the \ac{cnn}.
\\ \indent
It is notable that the parameter count decrease is higher than the decrease in filters. The reason is, that the higher the amount of filters in a layer $i$ the more likely is to encounter prunable filters. 
The amount of parameters stored in such layers are way larger, because the previous layers $i-1$ typically have higher filter counts as well. 
This behaviour is visualized for TinyYOLOv3 in Figure \ref{fig_pru2} by plotting the percentage of stored parameters per layer to the overall TinyYOLOv3 parameter count.
\\ \indent
With the reduction of the parameter count of approximately $23.1\%$ for \textsc{Frobenius} and $27.7\%$ for sparsity pruning a higher reduction of \acp{flop} can be observed compared to batchnorm fusing from Subsection \ref{sec_bn_benefit}. The comparison of \acp{flop} reduction percentages between original model (with batchnorm sub layers), fused model and pruned models is visualized in Figure \ref{fig_pru3}.
The reduction of \acp{flop} is $13.3\%$ for \textsc{Frobenius} and $15.7\%$ for sparsity pruning compared to the original model with batchnorm sub layers. 
\\ \indent
Conclusively, it has to be stated that a very high pruning result (e.g., $>80\%$ of parameters pruned) needs more investigation. The \ac{cnn} might either be oversized or not properly trained. In both cases, the filter weights are still close to the initialization. This leads to low values for the introduced metrics, which result in a huge amount of prunable filters.

\section{Quantization}
In this section, a method of changing the arithmetic backbone from floating point to integer arithmetic is given and evaluated with investigating on the deviation between floating point \ac{cnn} and integer \ac{cnn}. In conclusion, other methods are mentioned and briefly explained.
\label{sec_quant}
\subsection{Background}
Training for \acp{cnn} is traditionally performed using $32$ bit floating point arithmetic. The main reason is, that calculating the gradient during \ac{cnn} training requires high value resolution for which integer or fixed-point systems are insufficient. However, using floating point arithmetic on re-configurable hardware such as \acp{fpga} for \ac{cnn} inference comes with plenty of disadvantages: higher computation effort, more memory storage and increase of bus widths and counts. These disadvantages can be solved by using a quantized twin of the \ac{cnn} with a minimal or no drop-off in inference accuracy. 

\subsection{Quantization Approach}
A straight forward approach based on scaling the floating point values is proposed. 
This is realized by determining a positive whole-numbered scaling factor $S$, e.g., $S=256$. 
It is advisable to select this factor from the set of powers of two $S = 2^{P}$ where $P \in \mathbb{N}$, because integer division by a power of two becomes a right shift by the power of two $P$:
\begin{equation}
\frac{X}{S} = \frac{X}{2^P} = Rshift(X,P) \quad P \in \mathbb{N}
\label{eq_quant0}
\end{equation}
This is especially useful since the integer product $Z=MUL_S(X_1,X_2)$ of two mapped operands $X_i =x_i \cdot S$ requires a division by the scaling factor $S$ to obtain $Z = S\cdot x_1 \cdot x_2 + e = S \cdot z +e$ where $e$ denotes the quantization error:
\begin{equation}
\underbrace{Mul_S(X,Y)}_{=:Z} = \frac{X\cdot Y}{S} \stackrel{S=2^P}{=\joinrel=} Rshift(X\cdot Y,P)= S \cdot \underbrace{x \cdot y}_{=:z} + e
\end{equation}
Using shift operations instead of divisions spares hardware resources and reduces the combinatorial path, allowing a higher clock frequency of the hardware. 
With this background, the quantization approach can be introduced:
\begin{itemize}
\item \textbf{parameter and input quantization:}\\
Every parameter value $V$ (floating point, usually $32$ bit) of the \ac{cnn} is quantized by multiplying with the scaling factor $S$ and rounding to integer values. For this, the integer format $int16$ is used.
\begin{equation}
V_{quant} = int16(round(V*S))
\end{equation}
The input $A^{[0]}$ is quantized similarly for every pixel value $V \in A^{[0]}$.
\item \textbf{quantized convolution:}\\
The forward propagation of the convolution layer, which is split up into the filter convolution $Conv$ and the bias addition $Add$, is adjusted by appending a right shift operator $Rshift$ after the convolution step as illustrated in Figure \ref{fig_quant1}. 
\begin{figure}[htbp]
\centerline{\includegraphics[width=0.5\textwidth, trim = 0 0 0 0, clip, page=2]{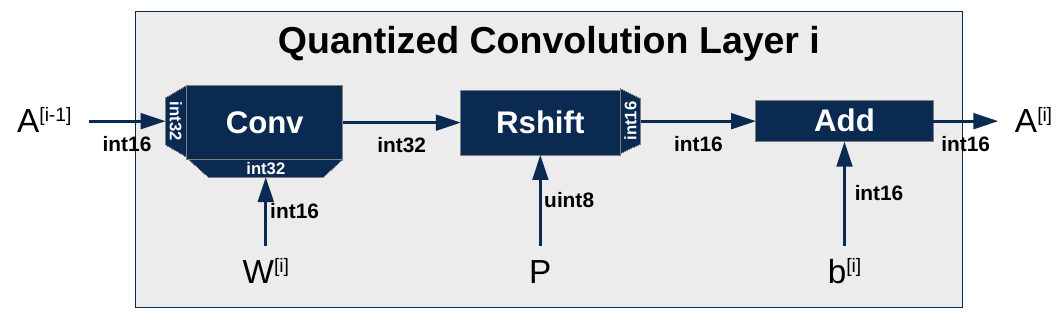}}
\caption{Quantized Convolution Layer $i$ without pooling and activation sub layer.}
\label{fig_quant1}
\end{figure}
The convolution itself is performed using $int32$ range with expansion of the previous activation $A^{[i-1]}$ and the filter parameters $W^{[i]}$ to $int32$:  
\begin{align*}
A^{[i-1]} &\leftarrow int32(A^{[i-1]})\\
W^{[i]}   &\leftarrow int32(W^{[i]})\\
A^{[i]}   &\leftarrow Conv(A^{[i-1]},W^{[i]})
\end{align*} 
After the right shift by $P$ the format is reduced to $int16$ and the bias $b^{[i]}$ is added:
\begin{align*}
A^{[i]}   &\leftarrow Rshift(A^{[i]},P)\\
A^{[i]}   &\leftarrow int16(A^{[i]})\\
A^{[i]}   &\leftarrow Add(A^{[i]},b^{[i]})
\end{align*}
\item \textbf{quantized activation function:}\\
An approved activation function for \acp{cnn} is the \ac{lrelu} as defined in \eqref{eq_quant1} and shown in Figure \ref{fig_lrelu}. 
\begin{equation}
a = ReLU_\alpha(z) = \left\{\begin{array}{ll} z, & z > 0 \\ \alpha \cdot z, & z \leq 0 \end{array}\right. 
\label{eq_quant1}                                                            
\end{equation}
The value of the subscripted $\alpha$ is denoting the negative slope.
Common values for $\alpha$ are $0.01$, $0.2$ and $0.3$ (default values of Caffe2, Tensorflow and Keras). 
For $\alpha=0$ the \ac{relu} without any \enquote{leakage} is obtained.\\ \indent
\begin{figure}[htbp]
\centerline{\includegraphics[width=0.4\textwidth, trim = 0 10 0 10, clip]{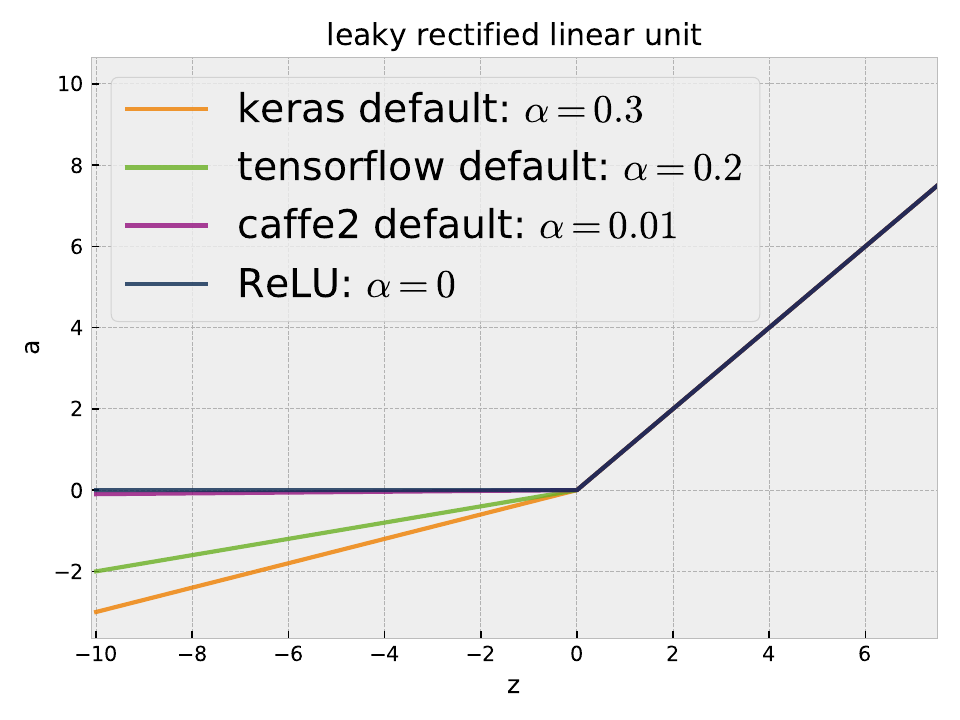}}
\caption{\ac{lrelu} for various $\alpha$ values.}
\label{fig_lrelu}
\end{figure}
\ac{lrelu} is a suitable activation function for targeting embedded systems, since its computational effort compared to other activation functions like hyperbolic tangent $tanh$ or sigmoid $\sigma$ is low. In order to make it even more suitable, the negative slope $\alpha$ is chosen from the set of negative powers of two: $\alpha = 2^{-P_\alpha}$ where $P_\alpha \in \mathbb{N}$: E.g., $\alpha=2^{-4}=0.0625$.\\ \indent
Analogously to the convolution quantization the multiplication by $\alpha$ is rewritten using the right shift operator $Rshift$ and the relationship from
\eqref{eq_quant0}:
\begin{equation}
\alpha \cdot z = 2^{-P_\alpha} \cdot z = \frac{z}{2^{P_\alpha}} = Rshift (z,P_\alpha)
\end{equation}
This simplifies \eqref{eq_quant1} for integer arithmetic as follows:
\begin{equation}
a = ReLU_\alpha(z) = \left\{\begin{array}{ll} z, & z > 0 \\ Rshift(z,P_\alpha), & z \leq 0 \end{array}\right. 
\label{eq_quant2}    
\end{equation}
\item \textbf{pooling:}\\
No adaptations are needed for transferring max pooling to integer arithmetic. For average pooling, further right shift simplifications can be made.
\end{itemize}
\subsection{Quantization Deviation}
\begin{figure}[htbp]
\centerline{\includegraphics[width=0.5\textwidth, trim = 10 10 10 10, clip]{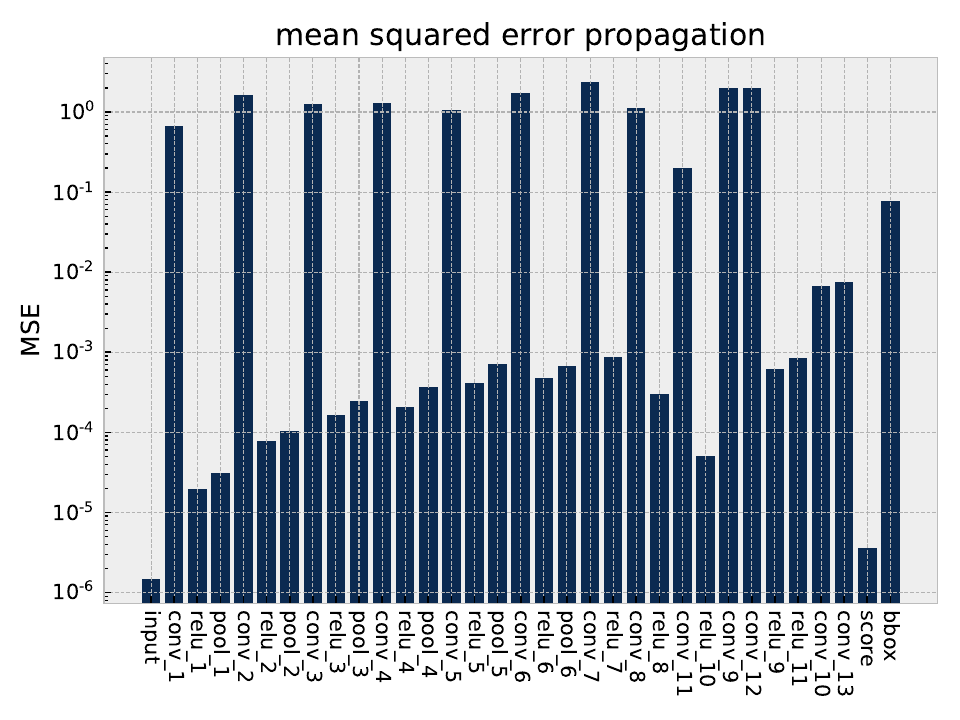}}
\caption{\ac{mse} (y-axis, logarithmic) per layer (x-axis) for TinyYOLOv3 using a scaling factor of $S=256$ on a randomly chosen image from the test set of TU Darmstadt Pedestrian Dataset \cite{p4}.}
\label{fig_quant2}
\end{figure}

The deviation between the trained floating point model and its quantized integer twin is estimated by the outputs of each convolution layer and the following sub layers (activation, pooling). The deviation is calculated using the \ac{mse} as metric:
\begin{equation}
MSE(A^{[i]}_{float},A^{[i]}_{int})= \frac{1}{N}\sum_{\substack{v \in A^{[i]}_{float} \\ w \in A^{[i]}_{int}}} \left(v-\frac{w}{S}\right)^2
\end{equation}
with $N$ denoting the number of elements (cardinality) of the compared arrays:
\begin{equation}
N=\mathcal{C}\left(A^{[i]}_{float}\right)=\mathcal{C}\left(A^{[i]}_{int}\right)
\end{equation} 
with $S$ being the scaling factor used for quantization.\\ \indent
Figure \ref{fig_quant2} shows the layer-wise calculated \ac{mse} between floating point model and integer model using a scaling factor $S=2^8=256$. 
The used network is the pruned version of the pedestrian detector network introduced in Subsection \ref{sec_pruning_results}.
 \\ \indent
A slow increase of the \ac{mse} is observable regarding the outputs of each TinyYOLOv3 layer. This increase plateaus around the fifth layer and stays below a \ac{mse} of $0.001$. The \ac{lrelu} activation function helps dealing with deviation from the convolution operation by dimming out negative deviation due to the lower negative slope. But most significantly is that the final convolution layers $conv\_10$ and $conv\_13$ display low \acp{mse}. This results in a neglectable difference at the detected classes scores $score$ of $\Delta_{score}=0.0019$ and a deviation of maximum $2$ pixel for the found \ac{bbox}.\\ \indent 
Similar behaviors with low or neglectable deviations are observed for other TinyYOLOv3 networks, which are trained for different detection tasks, as well as other architectures such as \ac{vgg} classification networks \cite{q0}.

\subsection{Advanced Approaches}
\label{adv_approaches}
In the following, other promising strategies and more advanced approaches are briefly described:\\[0.2em]  \indent
One approach explicitly tackles the size of the network compressing them by factors up to $30$ using weight sharing via k-means clustering method and code books with \textsc{Huffman} encoding \cite{q1}. \\ \indent
Another goal is to reduce the internal bit widths. In order to achieve that, a distribution based quantization scheme is developed quantizing each filter accordingly to the weight distribution. With that, lower bit widths like $4$ bit weights and $8$ bit activations are possible \cite{q2}. \\ \indent
Structural adaptations in architecture designs are proposed for example with MobileNet \cite{b3}. MobileNet introduces a new layer type: the depth-wise convolution layer. One of its advantages compared to the \enquote{classic} convolution layer is the lower amount of \acp{flop} required for inference. In MobileNet the \ac{relu} function is extended with a threshold parameter $\theta$ for preventing overflows and allowing lower bit widths. The default value used in MobileNet for $\theta$ is $6$. The so called \ac{relu}-6 function is directly supported in tensorflow \cite{t2.5}. \\ \indent
Another very promising idea is based on replacing multiplications with XNOR operations, which is speeding up inference heavily \cite{t2}. The so called XNOR-Net is created for image classification using binary activation and adapted gradient descent methods for training. However, its main disadvantage is the lack of framework support for the commonly used machine learning frameworks. \\[0.2em] \indent
In order to extend the existing design, appropriate concepts of the presented advanced approaches will be considered.

\section{Conclusion}
\label{sec_res}
The work flow of the adaptation stage with the following methods is shown:
\begin{itemize}
\item \textbf{Fusion of batch normalization sub layers} (Section \ref{sec_bn}):\\
Formulae and benefits for eliminating the batchnorm sub layers by fusing the batchnorm parameters into the weight and bias parameters of the preceding convolutional layer are presented.
\item \textbf{Pruning of convolutional filters} (Section \ref{sec_fp}):\\
A pruning routine with two different pruning metrics is elaborated. The advantages and reduction potentials are exemplary stated.
\item \textbf{Creation of a quantized twin using integer arithmetic} (Section \ref{sec_quant}):\\
A straight forward approach with utilisation of shift operation to speed up inference is given. More advanced strategies are mentioned and will be investigated in future works.
\end{itemize}
Overall, the goal of running a TinyYOLOv3 \ac{cnn} architecture on an Artix-7 \ac{fpga} is accomplished. In order to speed up inference on the \ac{fpga}, more optimizations will be developed and supplemented with fitting improvements from other strategies as presented in Subsection \ref{adv_approaches}.

\section*{Acknowledgment}
The investigation of the described adaptations for \acp{cnn} on re-configurable hardware are developed at AVI Systems at Freital Germany as well as the chair of VLSI-Design, Diagnostic and Architecture of the Institute of Computer Engineering at Technische Universit\"at Dresden.


\vspace{12pt}

\end{document}